# Five questions and answers about artificial intelligence

Alberto Prieto[†*] and Beatriz Prieto[†]

Department of Computer Engineering, Automation, and Robotics; CITIC-UGR, University of Granada, Spain

**Abstract.** Rapid advances in Artificial Intelligence (AI) are generating much controversy in society, often without scientific basis. As occurred the development of other emerging technologies, such as the introduction of electricity in the early 20th century, AI causes both fascination and fear. Following the advice of the philosopher R.W. Emerson's advice 'the knowledge is the antidote to fear', this paper seeks to contribute to the dissemination of knowledge about AI. To this end, it reflects on the following questions: the origins of AI, its possible future evolution, its ability to show feelings, the associated threats and dangers, and the concept of AI singularity

**Keywords:** Artificial Intelligence (AI), Fourth Industrial Revolution, Beginnings of AI, Development of AI, Automatic learning, Machine learning, Feelings in AI, Dangers of AI, Advantages of AI, Singularity of AI, Superintelligence, Frictionless Reproducibility (FR), Large Language Models, General AI (GAI), Intelligence, GPT Chat.

## 1  Introduction

Many researchers consider that since the end of the 18th century we have experienced several industrial revolutions [15]. The first, brought about by the invention of the steam engine, led to the development of the textile and metal industries. The second, known as the Technological Revolution, involved the advancement of the steel and oil industries, as well as the advent of electricity. The third is the Digital Revolution, which brought about the popularisation of computing and ICTs in general, with the introduction of personal computers and the Internet.

The concept of the 4th Industrial Revolution (4IR) was coined during the World Economic Forum (Davos Forum) in 2016 [16, 10, 21]. This expression refers to the ubiquity of technology, allowing ICT systems and devices to be available everywhere, accessible from anywhere and at any time. This latest revolution involves the full development of Data Science, Nanotechnology, Robotics, mobile ICT devices, the Internet of

---

[*] Corresponding Author  [†] Equal Contribution.

Things, Biotechnology, smart/autonomous vehicles, 3D printers and quantum computing. Artificial Intelligence (AI) is used in almost all these fields and is considered to be one of the pillars of the 4RI.

More and more information processing systems include AI algorithms, integrating into our daily lives in practically all contexts. Their potential to transform the foundations of our society is extraordinary, moving towards a progressive merging of the physical, biological and digital domains.

AI can be analysed from multiple perspectives, both scientific and non-scientific. This paper explores the following questions:

1. Can Artificial Intelligence have feelings?
2. How long has AI existed? Although it may seem new, the development of tools or machines capable of performing activities similar to those of the human mind is very old.
3. What will AI achieve? AI applications that until a few years ago seemed like science fiction are gradually being realised; how far can we get.
4. Which are the dangers of AI? Like any technology, AI can cause damages, we must claim and demand that developments truly contribute to progress, equality and prosperity for all.
5. What does it mean to reach the singularity of Artificial Intelligence? One of the futuristic visions of AI advances consists of developing a system capable of equalling and even surpassing the human mind. Will machines ever be able to surpass their own creators in intelligence? Will the realisation of a non-biological superintelligence be possible?

## 2. Can Artificial Intelligence have feelings?

In our relationship with other people, machines and elements in our environment, what really matters is what and how we perceive those sensations. We can judge and categorise people (friendly, unkind, immoral, etc.) but in practice, it is the particular interactions with them that concern us. Thus, we say that a person is empathic when he/she is able to identify with someone and share their feelings. This is a conclusion we draw from their behaviour, irrespective of the nature or essence of that individual.

From this point of view, it can be said that a machine can appear to have feelings if, in its interactions with people, we perceive attitudes similar to those of humans. Indeed, we can appropriately programme machines to behave as if they have feelings or emotions, in the sense that, in concrete situations, they emulate phenomena such as joy, affection, humour, love, happiness, fear, anxiety, anger, sadness, anger, rejection, shame, guilt, surprise, hope or compassion.

To illustrate the previous ideas, mention may be made of the lecture by Dr. Carme Torras, researcher at the Institute of Robotics and Industrial Informatics (CSIC-Polytechnic University of Catalonia), at the Spanish Congress of Informatics (CEDI2024) held recently in La Coruña, that brings together the main Spanish scientific societies in the field of Informatics. In this conference, Dr. Torras discussed "The multiple facets and challenges of assistive robotics". The robots that have been designed and programmed at her institute have met very demanding challenges to ensure reliable, personalized communication that respects the values and dignity of caregivers and patients, as well as guaranteeing the strict security measures that physical contact requires, for example, when helping users with mobility limitations to dress or feed themselves. These "social" robots have been designed in such a way that, in addition to meeting their basic objectives, they exquisitely maintain human-robot interaction. They do this by using the information that the robot collects and updates on the personality, tastes and specific behavioural progress of each patient, and by programming spoken or textual expressions that, according to the stored profile of the patient, show feelings such as affection, joy, or anger. They address each person by their name or nickname and say things like "Antonio, cheer up, you're doing very well, I'm sure we'll do better today", "Yolanda, you have to try a little harder because if others have done it, you will too", "Do you want some more water?", "I hope you have a nice day", etc. The machine does not understand these expressions at all, acting like a parrot, but it says them at the right time and putting the name of the person it is addressing before it, all in accordance with what has been programmed. They fake emotional empathy, which facilitates their autonomous behaviour, thus increasing their independence and adaptability in a socially dynamic context [1].

In another lecture, given by Dr. López de Mántaras, in which he reflected on the 'intelligence of artificial intelligence', he cited the case of presenting his students a group of robots moving randomly on a flat table. The robots moved without crashing into each other and when they approached the edges of the table they stopped and changed direction not to fall off. Upon observing this, one student immediately exclaimed 'they are afraid'. The student perceived and interpreted the robot as having the feeling of fear, which was only a behaviour programmed with the data obtained from the sensor readings of the table edge. If the programmer had set his mind to it, he could have made the robot ostensibly give the sensation of being afraid, by shouting, hesitating, turning from left to right, and so on.

Also, systems that we call intelligent can be endowed, for example, with what we call 'politically correct behaviour'. For example, if I ask Alexa: 'Alexa, tell me a joke about Andalusians', she might reply something like: 'Sorry, but I don't consider this request appropriate because it mocks a particular community'. This example shows that intelligent systems can be made to behave in an ethically correct way, or, to abuse the language, to say that they 'have ethics". To say that the machine 'behaves ethically' is a lot, but it is meaningless to say that it does or does not have ethics; there are countless unpredictable situations in which pre-programmed responses cannot be predicted.

From the above examples, car we infer that machines can have feelings? It should always be kept in mind that machines, and AI in particular, do nothing on their own. Machines only do what is intended by the people who conceive, design, program and implement them, and they can seem to have feelings, which in certain contexts can be beneficial. Machines can be programmed to fake feelings, but they will never have them. On the other hand, it is useful to consider, as López de Mántaras does, the 2020 statement by E. Bender and A. Koller: 'creating technology that emulates humans by pretending to be human requires us to be very clear about what it means to be human, otherwise we run the risk of dehumanising ourselves'. Although empathy can help technology better meet human needs, it can also be deceptive and potentially exploitative [3].

In conclusion, we can say that it is possible to perceive that machines have feelings and behave in an ethically correct way, but it makes no sense to say that the machine, in its essence, has them, unless we accept the phrase from Pirandello's famous play: "Right You Are-If You Think You Are".

## 3. How long has IA existed?

To speak properly about Artificial Intelligence (AI), first of all, we should be clear about the concept of 'intelligence'.

There are many definitions of intelligence, and even different types are considered (linguistic, emotional, logical-mathematical, musical, existential, creative, etc.). In the different meanings of the RAE (Royal Spanish Academy) intelligence is associated with the ability to understand, comprehend, know and solve problems.

AI is an application of computer science in the sense that it deals with problems related to automatic information processing. An AI system involves the collection of information (textual or from sensors), the representation and implementation of algorithms through programs, and the implementation (execution) of these algorithms on general-purpose or specific computer. The results can be the analysis of information for automatic decision making, or the generation of control signals on actuators such as those governing the movement of a robot. Algorithms describe the behaviour of the system but, by themselves, they do nothing (they are simply descriptions of tasks to be performed) and need a computer to execute them.

Among the various definitions of AI are John McCarthy, Marvin Minsky y Claude Shannon (1956) [13]: 'Artificial intelligence is the science that makes machines do things that would require intelligence if done by humans', or the RAE's: 'Scientific discipline concerned with creating computer programmes that perform operations comparable to those performed by the human mind, such as learning or logical reasoning'.

These definitions cite the human mind or intelligence. From our perspective, AI originated a long time ago, considering how the concept of intelligence has evolved over time.

Until a little over than a century ago, only extraordinarily intelligent people were considered to know how to multiply and divide, and these operations were considered genuine evidence of the human mind's capacities for understanding and problem solving. An anonymous manuscript from the Queen Elizabeth era (1570) contained sentences such as: 'Multiplication is a vexation', "Division is the evil", "The rule of three is a puzzle and by using it we go mads". In this context, it is clear that instruments such as the Pascaline, created to add and subtract by Blas Pascal in 1642, as an aid to his father a Normandy's tax collector, and later Gottfried W. Leibnitz's mechanical calculator (1672), which also performed multiplication and division, would have been considered artificial intelligence systems in their time.

In 1946, Eckert (an electrical engineer) and Mauchly (a physicist) presented the ENIAC at the University of Pennsylvania, which is considered the world's first electronic digital computer. In the report in which they describe the conception of this system, it is expressly stated that they try, to a certain extent, to emulate human behaviour in the sense that, just as people have organs to capture information from the outside (senses), they provide their system with input units, ..., and, as humans have a brain that controls and regulates the functioning of everything, they design a control unit that generates the signals controlling the operation of the entire system. In other words, the idea of imitating the functional structure of the biological information processing system to perform tasks similar to those produced by the human mind underlies the first general-purpose electronic computer. In fact, until well into the 1960s, computers were referred to as 'electronic brains' and what we now call 'Computer Science' was often referred to as 'Cybernetics'. From our opinion, all these systems, according to John McCarthy, Marvin Minsky and Claude Shannon, and the RAE's definitions, are 'Artificial Intelligence'. They were (and they are) machines that reason deductively quite well, executing programmes that codified algorithms (which are ultimately deduction mechanisms) and thus solved difficult problems. It should also be noted that these computers, in addition to arithmetic operations, performed logical operations and made decisions consisting of executing one or other instructions depending on certain arithmetic or logical conditions.

Later, well into the 1990s, as computers became popular, it was said that machines could not really have intelligence, and, by definition, an information processing task was only considered intelligent (human) when it could not be performed by a machine. In those years, it was incessantly commented that 'a computer would never surpass human intelligence, and proof of this is that it would never be able to beat an expert human at chess'. In other words, while in the Renaissance multiplying was considered a task illustrative of human intelligence, at the end of the 20th century it was the game of chess. The mould was broken when in May 1997 the Deep Blue computer was able to defeat the world chess champion Garry Kasparov.

Traditionally, in order to develop an intelligent system, the starting point is a precise description (algorithmized) of the human intelligent process to be implemented, which implies knowing exactly the mechanisms underlying this process. In short, it is a matter of projecting features of the real (world) space into a symbolic space. For example, the visualisation of colour images on computer or TV screens is based on the development of algorithms that have been possible thanks to the knowledge of the Physical Theory of Colour. This approach, traditional in the field of Computer Science, is called top-down, and includes, for example, Expert Systems.

There is another orientation that is much more interesting than the previous one, and can be considered the genuine engine of today's AI. It is the bottom-up approach, also called connectionist. In this case, the computer creates abstractions from sensory stimuli and other data. This approach emerged mainly from the artificial neural network concept, which includes machine learning.

The origin of the concept of the artificial neural network can be found in two very relevant contributions. The first one corresponds to McCulloch and Pitts who in 1943 introduced a formal model of the neuron, consisting of an electronic circuit in which each of its electrical input signals is multiplied by a certain factor (weight), after which all the weighted signals are added together to produce a value (excitation) that is applied to an activation function that provides the neuron's output [14]. In the simplest case, the activation function was step-like, so that if excitation exceeds a certain threshold, the neuron triggers (active output). The weighting weights emulate the behaviour of biological synapses, which collect signals from other neurons, the excitatory function the cell body (soma) and the output the axon. The second contribution is based on the concept that the basis of the behaviour of the nervous system lies in the interconnection of an extraordinary number of neurons (computational elements) and that the specific functions they perform depend on the values of their synaptic weights and their excitation thresholds (Hebb 1949) [5]. Based on this idea, different learning procedures were devised. For example, in 1957, Rosenblatt presented a (supervised) learning rule to establish the value of the weights and threshold for the McCulloch and Pitts neurons, in order to carry out a specific task [17]. These systems were called "perceptrons", and in them, the neural network is made up of several layers of neurons, in which the outputs of each neuron are interconnected to the inputs of all the neurons in the next layer.

Based on the above ideas, various models of neurons and learning rules emerged, giving rise to the concept of 'Machine Learning'. The most important idea underlying machine learning is that, unlike the top-down approach, it is not necessary to know the specific and inherent details of how to perform a given task in order to perform it. Simply, input patterns are presented to the system, the response is analysed and, depending on whether it is correct or not, the network parameters (weights and thresholds) are modified. Thus, after a suitable number of learning samples, the error in the output is reduced and may even reach zero. With this learning model, tasks such as pattern recognition and association, system identification and prediction can be performed.

In other neural network models, it is not even necessary to analyse the output in the learning phase in order to modify the network parameters if necessary. Only the learning inputs and the appropriate algorithm are used to self-organise the network. This type of learning, called unsupervised, for example, can automatically identify hidden factors or properties inherent in the data, detect clusters, or complete and reconstruct incomplete data.

Today, many software applications include AI programs, but this name is so flashy and media-friendly that it is being used as a claim, in a non-rigorous and abusive way, and almost as a synonym for program. Rigorously, a function or program should be considered AI only if it has been developed using machine learning or, knowledge or reasoning discovery techniques. In other words, not everything is AI.

## 3 How far can artificial intelligence go?

One of the most important milestones in AI is the concept of 'machine learning'. With this technique, a neural network can learn to perform a certain information processing task, without knowing the detailed procedures (algorithm) of how that process is done in the natural world. To 'teach' a neural network in an AI system to perform a given process, a learning phase is implemented in which training pattern values are presented to the network input one by one, and the network's parameters are progressively set to the input one by one. The network parameters are progressively rectified according to an algorithm so that the output error is minimised. Depending on the size of the network, the number of input patterns required for it to learn can be extremely high.

For example, in the case of Chat GPT3.5 175B, 175 billion parameters (neuron weights and thresholds) of a 96-layer deep network had to be tuned. For the pre-training, a corpus of 570 billion words was used, compiled from phrases obtained from the Internet, from sites such as Wikipedia (3 billion), Common Crawl platform (410 billion), etc. [12]. It should be noted that, in order to create the language model, it was not necessary to know and programme the processes with which humans write and express ourselves, but rather to use general machine learning algorithms.

Once the system has learned, it goes into a second phase called recognition or production. For example, in the case of natural language models, when faced with user queries, the system creates statistically probable combinations of text which, in order to present the results, imitate natural language based on its training data.

Actually, the intelligence and ingenuity are not in the algorithms and machines, but in the people, who conceive and design them. A sample of the talent of the designers is found in automatic chess-playing systems. One of the problems in teaching these programs is to have a sufficient set of training patterns, which in this case are established by describing the successive positions of the pieces on the board from a given time. DeepMind's AlphaZero chess program is the one that in 2017 beat the Stockfish program, which at the time was the best game in the world. The designers of the new program managed to do the training phase without the need to externally provide patterns, simply by having the program know the rules of chess, randomly generating patterns, and playing the program against itself millions of times during the four hours of 'training'. Obviously, the program itself automatically determined the outcome of each training pattern (win, lose or draw) in order to properly rectify the network parameters. In short, the system was given the ability to learn by itself.

To answer the question of how far AI can get, we need to be aware that AI performs information processing operations and, depending on the results, can order certain tasks to be performed automatically. Everything a computer system, and in particular an AI program, does is what its designers intended it to do. A machine on its own is not capable of performing any operation that has not been planned and planned by humans.

Machine learning systems are of great interest, but they have serious limitations that are difficult to overcome, and that is that their behaviour really depends on training patterns. Only abstracted knowledge can be obtained from these patterns, and it is not possible to go beyond them. Let us analyse this statement by considering the large natural language models, and following the reflections on the 'intelligence of artificial intelligence' made by Dr. López de Mántaras in a lecture given at the Spanish Computer Science Congress (CEDI2024). This researcher, citing work by Rao Kambhampati, stated that 'language models, rather than understanding and reasoning in a general way, approximately retrieve text patterns contained in training data'. Training data describes tasks or events that have occurred (factual), but there are many other events that are theoretically or imaginatively possible (counterfactual) and it is very unlikely that they, or very similar ones, are included in the training. In the same vein, he also cited an article by Zhaofeng Wu et al. [22], where the performance of GPT-4 was analysed by trying to solve a diverse set of counterfactual tasks. From their findings, it is concluded that the apparent reasoning capabilities of large language models depend substantially on approximating and reusing patterns found in their training data. According to López de Mántaras, 'the most important thing here is that these patterns must have appeared in the training data in order to solve the task successfully. Otherwise, large language models fail hard. What they do is much closer to reciting based on approximations (some call it regurgitating words with slight paraphrases) than to reasoning.

To clarify the above, we posed the following question to various language model chats: 'Maruja has N=2 brothers and M=4 sisters. How many sisters do Maruja's brothers have?' None of the three chats used gives the correct answer (5 sisters): GPT-3 indicates 4 sisters, Copilot 2 sisters, and Gimini 'I can't help you right now with answers about elections and political figures…'. We asked Gemini the same question, but changing Maruja for Alicia, and, after reasoning the answer, she gives the wrong result of 4 sisters. In all cases they reason, but they start from false premises. The inability to generalise is also shown by the fact that by changing the order between sisters and brothers the answers are different, but they are still wrong.

People solve these kinds of problems using common sense, and this is what language models lack. These models learn from texts, images and videos in computer repositories and not all realised and hypothetically realisable life experiences are to be found in them.

Another important issue is that the AI we have today is 'specific' in the sense that they are systems that can only do a specific job, such as translating, playing chess, recognising speech or images, and they do this better than humans. The aim is to arrive at 'General AI' (GAI), which ideally would be able to perform any intellectual task, in any domain, that a person could do. For this I agree with López de Mántaras' statement: "AI needs to have a global model of the world, both physical and social, in which common sense knowledge is included". To achieve common sense, as a preliminary step to General Intelligence, 'we would have to start by developing systems of representation of the most basic foundations of human knowledge as namely time, space, causality, elementary knowledge of physical objects and their interactions, as well as of human beings. This would require multi-sensory perception so that systems learn from all possible sources of information: interacting with the world, interacting with people, reading, watching videos, and so on. The architecture of AI systems should use robust reasoning techniques (deduction, induction, abduction, analogy, common sense) to solve problems in unforeseen, uncertain and changing situations. Knowledge is essential to be incremental so that the system is constantly learning and relating what is new to what it learned before'. López de Mántaras finally stated: 'When we hear about some spectacular and mediatic success of an AI programme solving a complex problem, we tend to generalise. This leads us to think that AI has virtually no limits. In reality, what generative AI systems, which are the most striking ones today, have is not intelligence but 'skills without understanding', performing very efficiently tasks such as generating images or convincing text, but understanding absolutely nothing about the nature of what they generate.

AI does not have intelligence in the human sense, it only learns and makes deductions. It lacks creativity and cannot hypothesise, speculate, discover new things on its own initiative, or automatically apply its skills to other areas, as we humans do. Computers reason, but they do not think. Without input data or information, the machine alone does not know what to do. To be intelligent like people, they would have to be intuitive and creative on their own.

## 4 What are the dangers of AI?

Artificial Intelligence (AI) like any other technology has its advantages and disadvantages. Among the advantages are the following:

- **Making more informed decisions and predictions.** AI is able to process and analyse large volumes of data much faster than humans, leading to more efficient management of resources, and the possibility of making decisions and predictions that are fairer, more objective, transparent and evidence-based, unaffected by conflicts of interest, bias, self-interest, self-serving, corruption, etc.

- **Human-to-human communication.** AI, and Computer Science in general, improves human-to-human communication through Social Networking and other applications such as speech recognition or automatic translation, making it easier to interact in different languages and contexts.

- **Automation in critical sectors.** AI is transforming key sectors such as healthcare, energy, transport or education. This is enabling significant advances in personalised and predictive medicine, smart energy management, autonomous vehicles and individualised education.

- **Content Generation.** Generative AI can create new content, such as art, music, text and product designs, with little or no human intervention. This is revolutionising content creation and affecting the creative, cultural and entertainment industries.

- **Mass personalisation.** Generative AI enables personalisation of products, services and experiences on a massive scale, transforming how businesses interact with consumers and satisfy their specific needs and preferences, including dependent care, personalised medicine, and customised commercial offerings.

- **Creation and commercialisation of new products.** Thanks to the ability of generative AI to create designs and models quickly, innovation and new product development can be accelerated. This reduces time to market and increases global competitiveness.

Among the real risks today, which are not speculation, are the following, which involve trade-offs associated with certain advantages:

- **Change in the labour market.** AI is taking over routine or dangerous jobs, which benefits human labour by focusing on specialised skills. This requires new training strategies and adaptation policies for professions that will change or disappear.

- **Fraudulent reclaiming of the AI concept.** On the one hand, the term AI is being propagandistically abused to successfully introduce new services or products that are not really AI. On the other hand, large technology companies are exaggerating, and even lying, about AI achievements in order to gain more investment and improve their projection on the stock markets.

- **Global balance of power.** The race for AI superiority may alter the global balance of power. Leading AI nations could set new standards in diplomacy, defence and international regulations, creating a new geopolitical order. Technological advances should drive progress, equality and prosperity for all of society, not just for the few.

- **Access to massive amounts of data.** Large social media platforms and networks, managed by people with no democratic control, can access and exploit data from billions of people. This allows them to infer detailed information about our lives and potentially manipulate our behaviour subliminally, getting to know us better than we know ourselves.

- **Governance and surveillance.** AI has the potential to transform the governance of societies, with applications ranging from intelligent surveillance to secure voting systems. As in other areas of life, there is a dilemma between security and human rights. There are issues of privacy violations, monitoring and mass surveillance of citizens.

- **Influencing information and public opinion.** AI is a very powerful tool for facilitating manipulation and polarisation, as it has the ability to create fake news, images and videos that can pass as authentic. It can also lead to other types of disinformation that can have a significant and damaging impact on public opinion, politics and elections. Impersonation and falsification of people's identities should be considered as a crime as serious as, or more serious than, for example, counterfeiting money.

- **Faulty or spiteful AI designs and implementations.** AI systems, like other computer systems, are built according to the standards set out in 'Software Engineering', which is a discipline that establishes methodologies to implement programs and applications that are efficient, safe and follow ethical principles. The principles of Software Engineering must be followed especially in sensitive or critical AI applications, since, as in any human action, errors may occur. On the other hand, it should not be forgotten that irresponsible people may exist who do not take extreme measures to realise secure systems without, for example, giving them autonomy to perform undesirable actions, or even using AI as a powerful tool for evil. Designers must protect their products from cyber-attacks and malicious use, and ensure, as far as possible, that computational decisions do not have negative consequences for people, nor being beyond human control.

- **Unbiased training.** Machine learning systems respond according to the information they learn from. By using huge amounts of data, systems learn with the same biases and tendencies as that data, and these biases are reproduced in the responses. For example, AI can automatically produce sexist and racist texts, if it has been trained on data with these biases. The problem could be solved by properly selecting the training corpus, which can often be difficult or even impossible. In any case, with machine learning, the problem (intentional or not) of biasing algorithms towards certain tendencies by using incomplete training patterns can be found.

- **Identification of responsibility.** The development of a computer system, and in particular AI, involves a large number of people at different stages: design, analysis, algorithm development, programming, implementation, testing, validation, production and maintenance. An arising major problem is identifying responsibilities and accountabilities for the automatic decisions that systems may make. Machines do nothing on their own, they do everything according to programmes made by humans, so the immediate and ultimate results of the possible decisions that are programmed must always be foreseen, which is not always easy.

There are other long-term risks, including those associated with achieving the so-called 'AI singularity' within which speculation would lead to a super-intelligence that could monitor and control everything. Our goal should be that AI systems operate transparently, are subject to human oversight, and can be assessed and certified by competent external authorities.

## 5 What does the arrival of AI singularity represent?

Among the long-term risks of Artificial Intelligence (AI) that have attracted most interest is those associated with achieving the so-called "AI singularity".

One of the challenges proposed by several companies and researchers is to achieve Artificial General Intelligence (AGI), which would mean achieving a hypothetical system that would equal or exceed the average human intelligence, being able to perform any intellectual task that people could execute. In particular, machines could improve their own designs in ways not envisaged by human engineers, and recursively optimise themselves by becoming, albeit gradually, more and more intelligent, leading to super-intelligence. At that hypothetical point, the 'AI singularity' would have been reached.

Donohoe, in a recent article [4], argues that the driving force behind the rapid development of AI lies in the fundamental changes in computational research over the last decade. We are witnessing a profound transition towards what can be called Frictionless Reproducibility (FR), based on the combination of three key principles of Data Science:

a) Datafication of everything. Publicly available datasets can now be found online in virtually every domain.

b) Code sharing. Software and artefacts are freely shared, allowing many different researchers to run exactly the same workflow.

c) Competitive challenges. Competitions at conferences and company calls for proposals to obtain the best results (in execution time and accuracy of results) by competitors in performing a given task on predetermined data.

The application of RF dramatically increases the rate of propagation of scientific ideas and practices, affects mindsets and erases memories of what has gone before. The time between remarkable discoveries is steadily reduced. With FR, the human effort needed to reproduce a computational result decreases, tending to zero. This leads to a decrease in the time for a field to globally adopt a new dominant methodology. Projecting into the future, we could reach a situation where these times are zero. At that point, according to Donoho, we would have reached the singularity of Data Science.

The concept of AGI has its disbelievers and its fanatics, giving rise to numerous science fiction writings and films. To some extent, a new religion is being created. Some people consider the Internet, and everything surrounding it, to be a new God with attributes such as omniscience (complete knowledge of things that have happened, happen and will happen), omnipresence (being present everywhere), truth (sum of all truths), invisibility, and omnipotence (power over all things). Some also believe that the supposed artificial superintelligence is near and that it will be able to monitor and control everyone. Even the most doomsayers admit that homo sapiens in its present form may be coming to an end.

To develop and manufacture a new machine or system it is necessary to follow steps such as: conception, analysis, design, procurement and transport of materials, assembly of components, testing, corrections and fine-tuning, installation, etc. It is only science fiction that all these phases, and the chaining between them, could be carried out globally and autonomously, without human intervention and control. To implement an IAG system would require enormous amounts of financial resources and time. Furthermore, we fully agree with the opinion of López de Mántaras presented at the CEDI2024 congress, who said that 'human intelligence is the main reference for achieving the ultimate goal of an AGI, but no matter how advanced this hypothetical AGI becomes, it will always be very different from human intelligence'. The basis of this idea lies in the fact that 'the mental development required for any complex intelligence depends on its interactions with the environment, and these interactions depend on the body, in particular the perceptual system and the motor system'. Different bodies, with different perceptual and motor systems, necessarily give rise to different minds. The body shapes the way we think. Even if we were to provide a humanoid robot with all kinds of sensors (sight, smell, taste, hearing and touch) and motor actuators to produce all kinds of movements, the body of this robot would never be the same as that of humans, since its constituent elements, its behaviour, its energy consumption, its maintenance, reproduction, etc. would still be different. For example, a natural neuron is a living element that reproduces itself or its connections in the areas where it is needed. It is made up of biological material and is governed by biochemical phenomena, with information being transmitted between neurons by means of electrical signals. In artificial neurons, information is also transmitted by electrical signals, but the constituent material is silicon or gallium arsenide (integrated circuits).

Futurist and Director of Engineering at Google, Raymond Kurzwei [7, 8], who predicted that the AI singularity will arrive in 2045 [9], and other scientists [19, 6, 18, 11] conjecture that it is possible to build a complete brain. This could be done with artificial neural networks. However, it should be borne in mind that the brain has about 86 billion neurons, each of which can have about 10,000 synaptic connections to other neurons, with 100 trillion interconnections in total. Realising this structure, even though computer simulations, is extremely complex. Furthermore, this system would have to pass through learning phases in all domains of knowledge using appropriate perceptual inputs and motor outputs.

In any case, if we were to protect ourselves from the possible effects of the hypothetical arrival of an AGI, and within it a super-intelligence, we would have to ask ourselves philosophical and sociological questions.

However, it must be emphasised that machines only do what we humans have designed them to do. It is not the machines or the algorithms they run that are responsible, but the people who have conceived and implemented them. AI systems must be transparent, under human supervision, and assessable and certifiable by external authorities. It is essential to ensure that the data used to train these systems are free of bias and that fundamental rights are always respected. Proper legislation must exist, and, most difficult of all, monitoring systems to ensure that they are enforced. Fortunately, several bodies (UNESCO, European Union, different states, etc.) are undertaking this work. Let us hope that these rules will prosper without limiting their development.

## 6. Conclusions

AI is emerging as the main focus of the sixth industrial revolution and a key catalyst in the emergence of a new world order. This paper reflects on several crucial aspects, such as the origins of AI, its possible future evolution, its ability to simulate feelings, the associated threats and dangers, and the concept of AI singularity. The main conclusions are presented below.

Defining AI as the discipline that makes machines perform tasks that, if done by humans, would require intelligence, we can place its origins in the Early Modern Era, when the first calculating machines were created.

In terms of future developments, some of the challenges would be to develop machines capable of emulating human mental capacities, such as reasoning, comprehension, imagination, perception, recognition, creativity and emotions. Although we are still far from achieving these goals, very significant partial progress has been made. On the other hand, while AI systems can simulate emotions in a useful way in certain contexts, they are not capable of experiencing real feelings.

Among the current, non-speculative problems related to AI, several major drawbacks stand out. One of the most discussed is the destruction of jobs, requiring the development of new training and adaptation strategies. In addition, there exists a propagandistic misuse of the term 'AI', attributing it to systems that actually do not comply with its characteristics. Another relevant problem is the global monitoring and control of data, which makes it possible to extract confidential information and even alter the balance of power at the global level, given that information is a strategic resource. This is compounded by the creation and dissemination of fake news with the appearance of authenticity, impersonation, and the increasing difficulty of tracing its origin and assigning responsibility in these processes.

Significant progress has been made towards the creation of an artificial general intelligence (AGI), i.e. an AI with a flexible intellect comparable to humans, thanks to the development of OpenAI's GPT-4 language model, which allows machines to converse with each other as if they were human beings [20]. This advance represents an approach to one of the biggest potential risks: the feared singularity of AI. Speculatively, it is believed that at this point a superintelligence capable of monitoring and controlling all aspects of reality could be achieved.

Despite the above dangers, it is essential to remember that everything a computer system, and especially an AI programme, does is the result of what its designers intended. A machine, on its own, cannot execute any operation that has not been planned and anticipated by human beings. In fact, intelligence and ingenuity reside not in the algorithms and machines, but in the people, who conceive and develop them.

AI does not possess intelligence in the human sense; it is limited to learning and making deductions. It lacks creativity and is not able to hypothesise, speculate, make discoveries on its own initiative, or automatically apply its abilities to different areas, as we humans do. Computers can reason, but they do not think; if they are not given initial data or information, they do not know what to do on their own. To reach a human-like level of intelligence, machines would have to be autonomously intuitive and creative.

AI systems must be transparent, subject to human oversight, and assessable and certifiable by external authorities. It is essential to ensure that the data used to train these systems is free of bias and that fundamental rights are always respected. The key challenge is to achieve developments that truly drive progress, equality and prosperity for all, not just for the few. To this end, it is crucial to have adequate legislation and, most difficult of all, to establish effective monitoring systems to ensure compliance. Fortunately, various entities such as UNESCO, the European Union and several states are already working in this direction.

## Acknowledgments

This work was partially supported by the Spanish Ministry of Science, Innovation, and Universities (MICIU) together with the European Regional Development Fund (ERDF/EU), through project PID2022-137461NB-C31 (funded by MICIU/AEI/10.13039/501100011033 and by ERDF/EU).

16    A. Prieto et B. Prieto